# HOPPR

# Medical-Grade Platform for Medical Imaging AI

White Paper

Kalina P. Slavkova, PhD
Melanie Traughber, DSc
Oliver Chen, MD
Robert Bakos
Shayna Goldstein
Dan Harms
Bradley J. Erickson, MD, PhD
Khan M. Siddiqui, MD

**Abstract**

Technological advances in artificial intelligence (AI) have enabled the development of large vision language models (LVLMs) that are trained on millions of paired image and text samples. Subsequent research efforts have demonstrated great potential of LVLMs to achieve high performance in medical imaging use cases (e.g., radiology report generation), but there remain barriers that hinder the ability to deploy these solutions broadly. These include the cost of extensive computational requirements for developing large scale models, expertise in the development of sophisticated AI models, and the difficulty in accessing substantially large, high-quality datasets that adequately represent the population in which the LVLM solution is to be deployed. The HOPPR Medical-Grade Platform addresses these barriers by providing powerful computational infrastructure, a suite of foundation models on top of which developers can fine-tune for their specific use cases, and a robust quality management system that sets a standard for evaluating fine-tuned models for deployment in clinical settings. The HOPPR Platform has access to millions of imaging studies and text reports sourced from hundreds of imaging centers from diverse populations to pretrain foundation models and enable use case-specific cohorts for fine-tuning. All data are deidentified and securely stored for HIPAA compliance. Additionally, developers can securely host models on the HOPPR platform and access them via an API to make inferences using these models within established clinical workflows. With the Medical-Grade Platform, HOPPR's mission is to expedite the deployment of LVLM solutions for medical imaging and ultimately optimize radiologists' workflows and meet the growing demands of the field.

## Introduction

Artificial Intelligence (AI) is revolutionizing healthcare, particularly in medical imaging, where it has demonstrated immense potential to enhance diagnostic accuracy, reduce clinical workload, and improve patient outcomes[1,2]. Radiology is a data-intensive field that relies heavily on the interpretation of complex images while incorporating information from previous reports and patient medical records. The complexity and redundancy inherent in radiologists' tasks have catalyzed significant AI-driven innovations. The integration of AI in radiology extends beyond traditional image classification tasks and is expanding into more sophisticated applications, including report generation and evaluation, predictive diagnostics, and functioning as a second reader[1,3]. These developments are largely fueled by the evolution of generative AI and foundation models, which are multimodal, large-scale models capable of understanding and processing diverse data types. Such generative models, trained on vast amounts of imaging and clinical data, can be fine-tuned to adapt to specific clinical applications while ensuring high levels of accuracy and reliability.

HOPPR's Medical-Grade Platform exemplifies these advancements by providing a comprehensive infrastructure for the training, fine-tuning, and deployment of AI models that adhere to regulatory requirements within clinical settings. Specifically, medical-grade status is conferred by the development of the HOPPR platform under a robust quality management system (QMS), following ISO 13485, that aligns with the stringent demands of healthcare oversight and sets a new standard for foundation models in medical imaging applications. The platform comprises four essential pillars—Platform, Data, Foundation Models, and Validation & Regulatory—each designed to address critical aspects of medical AI implementation. Its medical-grade hosting environment allows for the seamless integration of AI into clinical workflows, while its extensive multimodal data repository supports the training and validation of complex models. Furthermore, by incorporating fine-tuning capabilities and a robust validation framework, HOPPR's platform addresses key regulatory and ethical concerns, such as bias mitigation and traceability.

**Background**

*What is a Foundation Model?*

Foundation models are large-scale AI models consisting of billions of parameters, trained on extensive and diverse datasets—often comprising millions of examples—through scaled-up self-supervision[4]. This enables them to learn generalized features from the training data. Training foundation models necessitates vast computational resources and substantial amounts of high-quality data. Once trained, however, foundation models can significantly expedite the development of task-specific models through fine-tuning (See Figure 1). Fine-tuning refers to the process of taking a foundation model pretrained on large datasets and continuing to train on a smaller, curated dataset to improve performance for a specific task.

Previously, deep learning research predominantly focused on training smaller models from scratch—starting with randomly initialized weights or weights from a different task—for specific use cases. This approach resulted in task-specific models limited to the tasks they were trained to perform. Foundation models are shifting the AI research paradigm away from training models from scratch[5]. This shift allows developers to begin with a foundation model that has learned generalized data representations, enabling fine-tuning for specific tasks to be accomplished more rapidly and with significantly less data and computational resources compared to the traditional approach. For example, commercial large language models (LLMs), such as GPT-4 from OpenAI, are initially pretrained on vast amounts of publicly available text data to learn representations of text in various contexts.

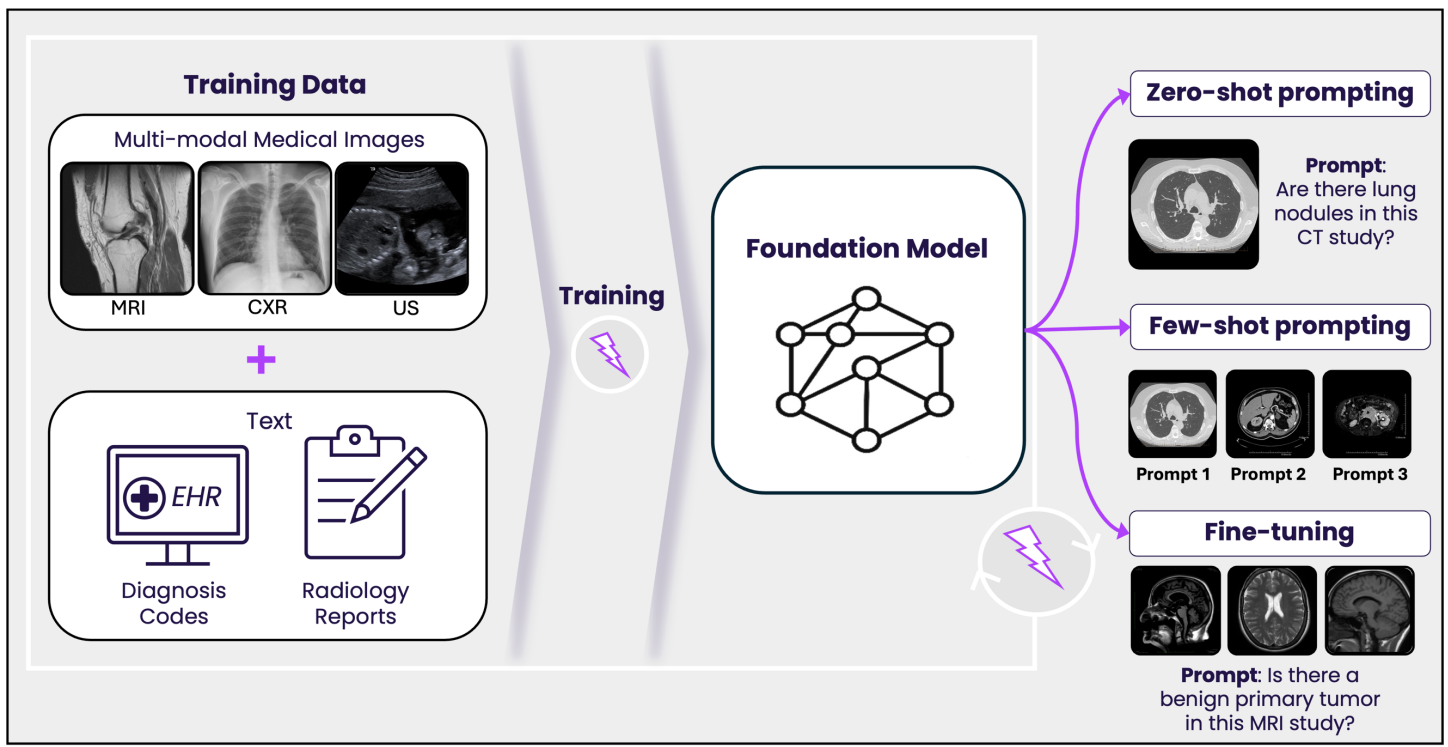

*Figure 1. Overview of a foundation model in the context of medical imaging.* Extensive amounts of data are curated for training, consisting of pairs of multi-modal medical images (e.g., MRI, chest X-Ray, and ultrasound) and text (e.g., EHR and/or radiology reports). Through training on many diverse examples in an unsupervised regime, the foundation model learns a generalized relationship between medical images and their clinical interpretations. Once the foundation model is finalized, it can be utilized in three broad ways: prompted in a zero-shot manner to complete new, unforeseen tasks (e.g., evaluating a lung CT when the training data did not include CT scans), prompted in a few-shot way to learn new information from a series of unforeseen prompts (e.g., the model gains knowledge through a series of prompts related to unforeseen data), and fine-tuned on smaller, curated datasets in which the model becomes better at a particular class of tasks (e.g., trained further on a small brain MRI dataset to become better at evaluating benign primary tumors).

Subsequently, researchers perform fine-tuning to align the base LLM with human values, resulting in a chatbot (e.g., GPT-4o from OpenAI) capable of conversing with human users.

The rise of foundation models has been facilitated by significant developments in deep learning research. One such advancement is the transformer architecture[6], upon which LLMs are built, enabling models to learn dependencies between different units of language (tokens) without sequential processing constraints. The Vision Transformer (ViT) applies the same concept of the original transformer to images, tokenized as small square patches[7]. The introduction of the ViT paved the way for the development of large vision models (LVMs) that can learn generalized relationships among tokenized images during pretraining. These technological developments have subsequently led to the popularization of vision-language models (VLMs)[8,9]. VLMs are trained on paired image and text samples and can be fine-tuned to perform tasks such as image captioning, image classification, and visual question answering, where the model can answer questions about an image. The latter is particularly relevant to VLM applications in medical imaging.

*Foundation Models in Medical Imaging*

When VLMs are scaled up to perform across more complex tasks, they are referred to as large VLMs (LVLMs). The application of LVLMs to medical imaging data has demonstrated significant potential in optimizing radiologists' workflows by automating time-intensive tasks, such as report generation and serving as a second reader for evaluating studies. Due to the high prevalence of chest X-ray and abdominal CT studies compared to other modalities like MRI, considerable efforts have focused on developing LVLMs for CT and X-ray processing. "Merlin" is one such VLM that has been pretrained on approximately 6 million images sourced from roughly 15,000 CT scans with corresponding diagnosis codes and radiology reports. It has been extensively evaluated on numerous clinical tasks, ranging from five-year disease prediction to report generation[10].

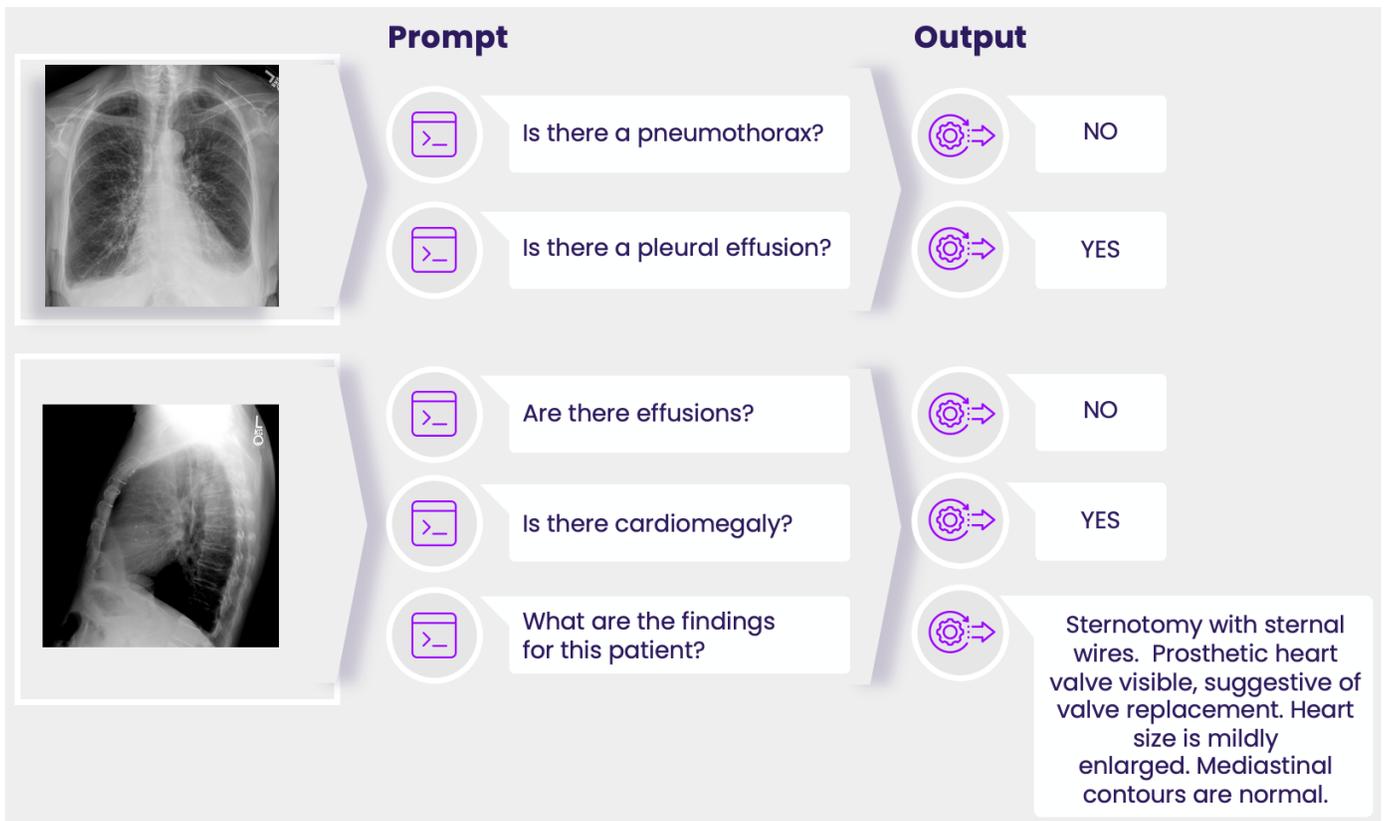

*Figure 2. Example prompts provided to the HOPPR Foundation Models.* The model is prompted with various questions assessing the presence of common findings on chest X-ray images, such as pneumothorax, pleural effusion, and cardiomegaly. The model can also be prompted with a descriptive prompt to detail all findings in the images.

Despite technological advancements and data availability enabling the use of LVLMs to improve radiology practice, several limitations hinder the deployment of this technology. As previously mentioned, training foundation models necessitates extensive computational resources and expertise as well as access to large, high-quality datasets from multiple imaging sites to ensure adequate population representation. Typically, isolated efforts to train foundation models for medical imaging applications are constrained by limited data sourcing from a small number of imaging sites, thereby restricting the demographic representation of the final pretraining dataset. Without demographic representation, foundation models cannot be performant across all populations.

To address this need, HOPPR has developed a platform maintained by a rigorous quality management system to streamline the adaptation process of LVLMs for clinical implementation that are performant across all socioeconomic and demographic populations. Additionally, there is a need to democratize foundation models to ensure they are trained with all beneficiaries in mind, particularly those in underrepresented, marginalized, and low-income settings, and a robust development process is fundamental to meeting this need.

**The HOPPR Medical-Grade Platform**

The HOPPR Platform is powered by proprietary LVLMs pretrained on the largest medical imaging dataset assembled to date, comprising over 120 million imaging studies spanning 10 years with 70 million added annually. This extensive dataset has been acquired through partnerships with over 400 imaging centers across eight states, encompassing all demographics and ethnicities and covering thousands of diagnostic assessments. A large, representative dataset is essential for developing models that perform robustly across all socioeconomic and demographic scenarios.

*A Platform for Deep Interaction with Imaging Studies*

The HOPPR platform offers model hosting capabilities for customers seeking to deploy custom models for evaluating imaging studies (see Figure 2 for an example use case). Through this partnership, users can upload studies and analyze them using their provided model via the medical-grade, secure HOPPR API that is hosted by a trusted cloud service and managed by a dedicated platform team. This API-based solution allows customers to seamlessly embed the capabilities of the HOPPR Foundation Models within their existing applications and workflow.

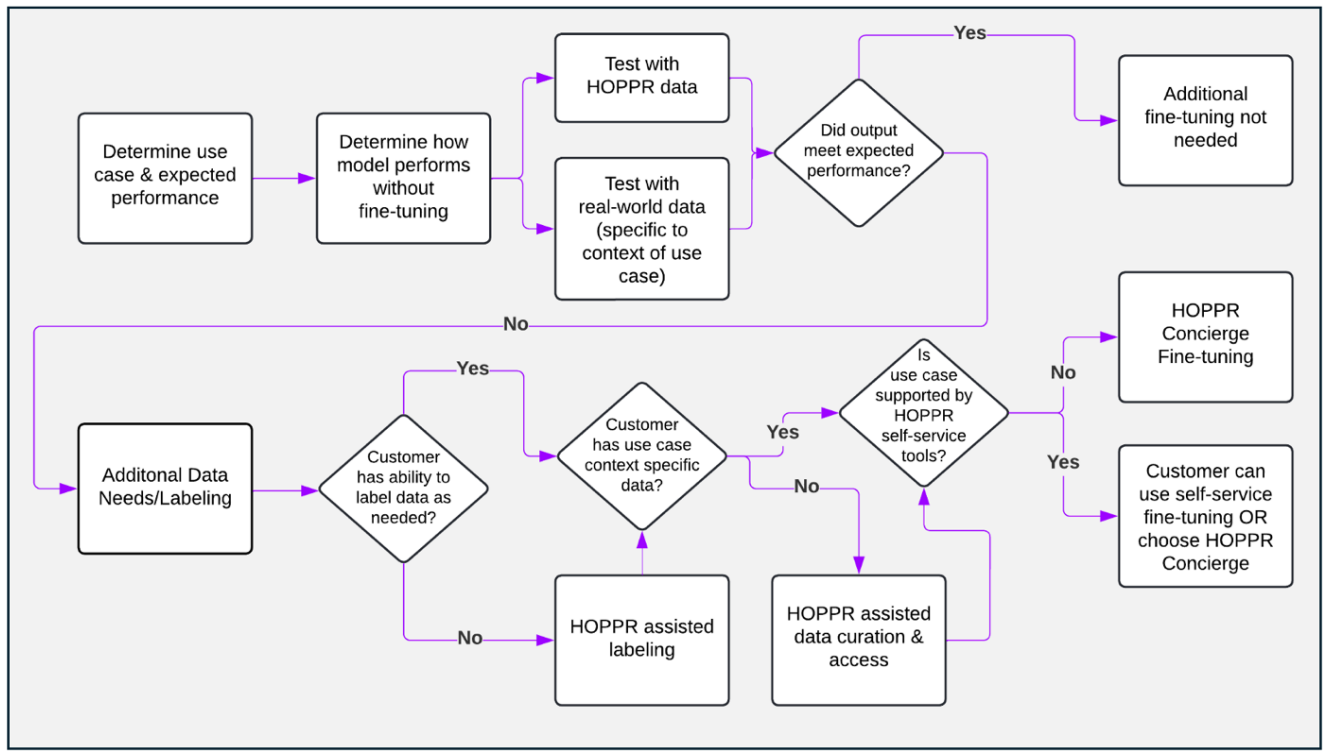

*Figure 3. Schematic demonstrating how HOPPR meets customer needs.*

*Fine-Tuning for Customer Use Cases*

To promote AI democratization, HOPPR provides powerful computational resources and expertise for developing custom models based on the HOPPR Foundation Models (see Figure 3). By fine-tuning these foundation models, customers can create custom models in a significantly shorter time frame with substantially less data and computational requirements compared to training task-specific models from scratch. The typical workflow for fine-tuning with HOPPR's support involves the following steps:

1. **Initial Evaluation**: For a given customer use case and required performance level, the initial step involves evaluating the HOPPR Foundation Models out-of-the-box on the task (i.e., without fine-tuning) using HOPPR data as well as real-world data specific to the context of the use case. If performance at this stage is sufficient, no fine-tuning is necessary.
2. **Data Acquisition and Curation**: If the expected performance is not achieved, the next step is to acquire additional data that fits the context of the use case. If the customer cannot provide this data, they may opt for HOPPR-assisted data curation and labeling for the specific use case.
3. **Fine-Tuning**: HOPPR plans to provide and expand self-service tools for fine-tuning that customers may use if the use case is compatible with the available tooling. Otherwise, HOPPR experts will develop a custom fine-tuning solution for the customer.

*How is Data Security and Privacy Ensured?*

Given the vast amount of data that HOPPR has obtained through site partnerships, it is imperative that a secure, HIPAA-compliant system is in place for data management. HOPPR de-identifies all data by transforming entries in the DICOM headers. For example, patient age is recorded as 89+ years, and the acquisition date is randomly shifted by ±7 days. Additionally, HOPPR employs "hiding in plain sight"[11] and many other techniques to ensure there is no malicious or accidental disclosure of Protected Health Information (PHI). A proprietary process is implemented to remove any "burnt in" PHI in the images as well. These transformations ensure that no PHI data remain in the reports and images.

Furthermore, HOPPR utilizes a trusted and fully managed cloud storage platform that encrypts and compresses medical imaging data and reports to ensure secure storage. HOPPR maintains stringent contracts with data partners that prohibit the sharing of data beyond the HOPPR ecosystem. Additionally, data provided by a customer to fine-tune a custom model will not be ingested into the HOPPR platform or used in the HOPPR Foundation Models in any way.

**Conclusion**

The HOPPR Medical-Grade Platform provides the infrastructure and a suite of tools to expedite the development and deployment of AI-based solutions for improving radiologists' workflows in an era of ever-growing demand for medical imaging. A key component of the HOPPR platform, the HOPPR Foundation Models, are large vision-language models pretrained on large amounts of high-quality imaging studies with corresponding text reports. Customers can fine-tune these models with additional context-specific data to improve performance for specific use cases. With this platform, HOPPR anticipates increased development of state-of-the-art AI tools and ease of clinical implementation that will address the current and future needs of radiology in all communities.